%% file: main.tex
\pdfoutput=1

\documentclass[11pt]{article}

\usepackage[preprint]{acl}

\usepackage{graphicx} 
\usepackage{natbib}  
\usepackage{caption} 
\usepackage{amsmath}
\usepackage{times}  
\usepackage{helvet}  
\usepackage{courier}  
\usepackage{graphicx} 
\usepackage{booktabs}
\usepackage{amssymb}
\usepackage{CJKutf8}
\usepackage{tabularx}
\usepackage{array}
\usepackage{algorithm}
\usepackage{algorithmic}
\usepackage{amsmath}
\usepackage{newfloat}
\usepackage{listings}
\title{TokenPrint: A Calibrated Token-Space Fingerprint for Language-Model Provenance}

\author{
Yuqi Wu$^{1}$,  
Shengming Zhao$^{1}$,
\textbf{Jie Chen}$^{1}$\thanks{Correspondence, jc65\@ualberta.ca} \\
$^{1}$College of Biomedical Engineering, Fudan University
}

\begin{document}
\maketitle
\begin{abstract}
Establishing the provenance of a language model---including its base checkpoint and possible overlap in training distributions---is a governance challenge that metadata alone cannot resolve. We introduce a training-free fingerprint based on the top-$k$ vocabulary projections of late hidden states elicited by 250 fixed knowledge probes, compared using Jaccard overlap over decoded token strings. We evaluate the method on 32 open-weight models from nine families (0.6B--32B) with documented relationships. (1)~A \emph{similarity ladder} broadly follows model relatedness: independently trained models on identical data score 0.48 raw (0.35 vocabulary-corrected), followed by shared-base fine-tunes (0.39/0.33), same-developer relatives (0.38/0.28), and models with no documented relationship (0.22/0.17). This identical-data signal persists across three organizations, two tokenizer families, and two architecture classes, and emerges within the first 1\% of training before measurable task competence, suggesting a contribution from shared training data beyond capability convergence. (2)~As a nearest-neighbor \emph{lineage-retrieval} method, the fingerprint ranks the exact documented base among the top two candidates for all five R1 distillations (mean rank 1.8, MRR 0.60), including a math-specialized base not identifiable from coarse metadata. (3)~A \emph{depth ablation} shows that lineage group discrimination strengthens toward the output distribution, with AUC increasing from 0.72 at quarter depth to 0.90 at the output; using only the top 5 output tokens retains AUC 0.87. (4)~The fingerprint remains stable under quantization, with Jaccard similarity of 0.92 under int8 and 0.82--0.85 under int4, compared with a maximum cross-model similarity of 0.81 in the calibration pool. We release the probes, code, and fingerprints.
\end{abstract}

\section{Introduction}

Open-weight language models are released faster than their provenance can be audited. Model cards may omit the base checkpoint used for fine-tuning, training-data disclosures are often incomplete, and nominally independent models may share substantial portions of their corpora. Existing forensic methods address parts of this problem. Weight-space approaches recover fine-tuning trees when parameters are shared \citep{horwitz2024origin,zeng2024huref,tong2026seedprints,wu2025gradient}, while representation-based methods such as REEF test whether a suspect model derives from a specific source model \citep{zhang2025reef}. These methods assume parameter inheritance. Behavioral phylogenies such as PhyloLM \citep{yax2025phylolm} apply to arbitrary model pairs, but yield uncalibrated structure: a dendrogram reveals which models cluster, not what level of similarity should be expected from a shared base, overlapping training data, or similar task competence.

We study a simple training-free fingerprint: the top-$k$ vocabulary projections of late hidden states on 250 fixed knowledge probes, compared by Jaccard overlap of decoded token strings. We evaluate it on a 32-model pool with documented relationships spanning multiple levels of relatedness: Pythia models \citep{biderman2023pythia}, which share training data, order, and recipe but not weights; DeepSeek-R1 distillations \citep{deepseek2025r1}, which inherit published base checkpoints; same-lineage-group models; successive generations from the same developer; and models with no documented relationship. An extended witness pool adds models trained on the Pile by three organizations, with different tokenizers (Cerebras-GPT \citep{dey2023cerebras}), codebases (GPT-NeoX-20B \citep{black2022gptneox}), and architecture classes (RWKV-4-Pile \citep{peng2023rwkv}).

\begin{figure*}[!htbp]
\centering
\includegraphics[width=0.8\linewidth]{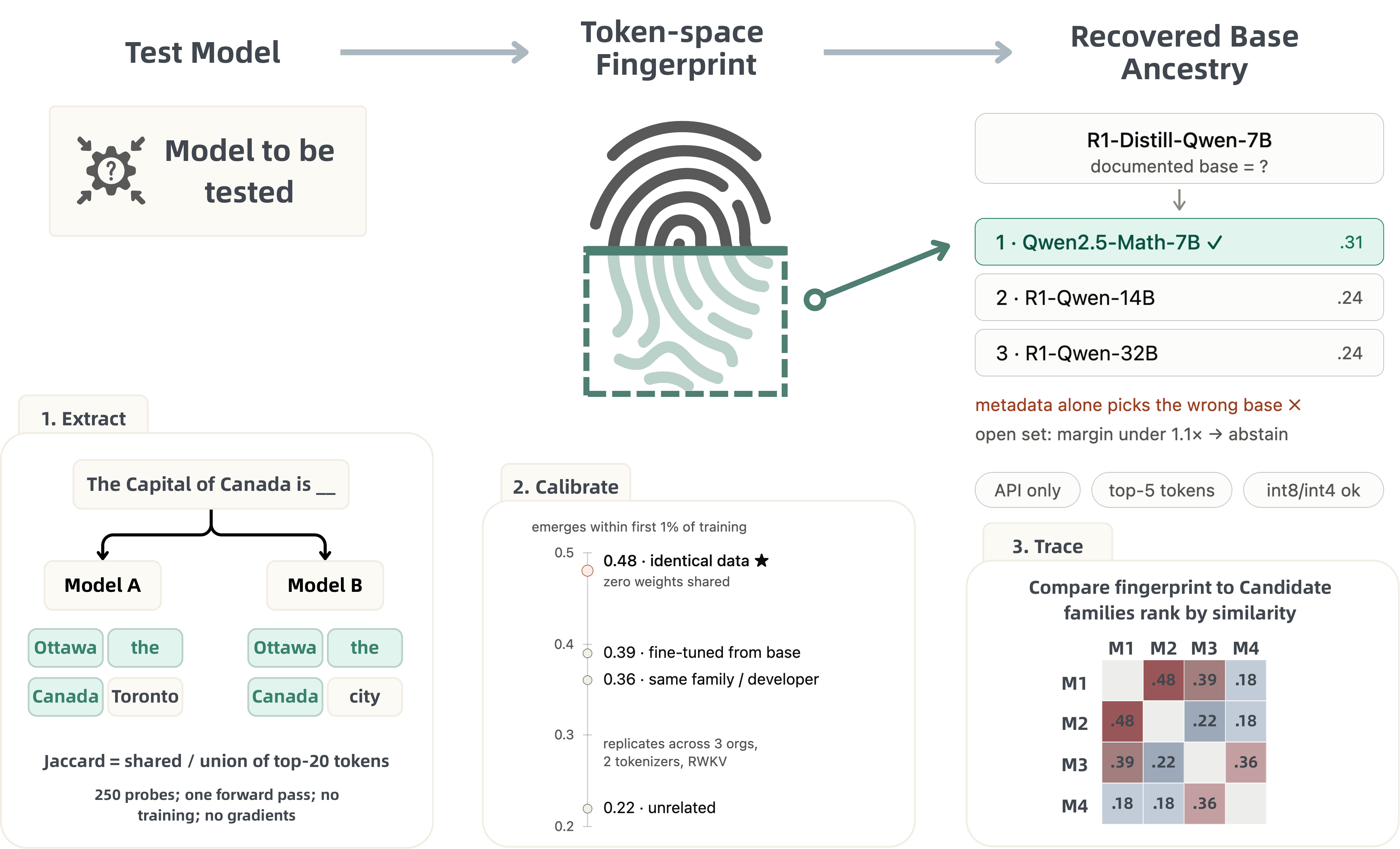}
\caption{Overview: each model answers 250 fixed knowledge probes; top-$k$ token sets at matched depths are compared by Jaccard overlap; pairwise scores are calibrated against documented model relationships.}
\label{fig:graph_abs}
\end{figure*}

Calibration yields three main results. First, fingerprint similarity follows documented model relatedness and remains informative across organizations, tokenizers, and architectures. Second, nearest-neighbor retrieval consistently narrows each R1 distillation to a small lineage neighborhood, while also exposing the difficulty of distinguishing an exact parent from closely related alternatives using output behavior alone. Third, the signal becomes stronger toward the output distribution and remains usable under restricted top-$k$ access and model quantization. Together, these results characterize both the utility and the limitations of token-space fingerprints for model-provenance analysis.

Overall, this work introduces a training-free token-space fingerprint with an explicit shared-vocabulary null, evaluates it across documented model relationships, and studies its behavior across training stages, model depth, access constraints, and quantization. It further shows how the resulting similarity structure can support lineage retrieval while avoiding claims of exact provenance that output-space evidence cannot reliably establish.

\section{Related Work}

\textbf{Weight- and representation-space fingerprints.}
HuRef derives human-readable invariants from parameter matrices \citep{zeng2024huref}; Model Tree Heritage Recovery reconstructs fine-tuning trees from weights \citep{horwitz2024origin}; SeedPrints traces lineage to random initialization \citep{tong2026seedprints}; gradient-based features support lineage group classification \citep{wu2025gradient}; and REEF compares paired activations using CKA \citep{kornblith2019similarity} to test derivation hypotheses \citep{zhang2025reef}. These methods require parameter access and primarily target settings with parameter inheritance. Our fingerprint requires only forward passes and also applies when models share no weights.

\textbf{Behavioral phylogenies and output analysis.}
PhyloLM infers phylogenies from sampled-output similarity across a large model pool \citep{yax2025phylolm}, while refusal directions have been proposed as behavioral fingerprints \citep{sun2026refusal}. Our output-level fingerprint belongs to this broader class, but focuses on calibrating pairwise similarity against documented model relationships. We also compare against a probability-overlap baseline based on the PhyloLM distance under the same model pool and evaluation protocol. Layerwise logit trajectories have been used to detect benchmark contamination within individual models \citep{zhu2025logittrace}; we use related readouts for cross-model comparison.

\textbf{Distillation and training-data detection.}
Membership and dataset-inference methods test whether particular examples or datasets were used in training \citep{shi2024detecting,zhang2025distilldata}, while distillation detectors test specific teacher--student hypotheses \citep{shi2025kddetection}. Watermark persistence has also been studied as a deterrent to unauthorized distillation \citep{pan2025watermarkkd}. These methods typically address instance-level or pair-specific questions. Our method is complementary: it produces a global similarity structure from 250 forward passes per model, which can be used to generate lineage hypotheses for subsequent targeted analysis.

\section{Method}

\subsection{Fingerprint Extraction}

Given a causal language model with $L$ layers, unembedding matrix $W_U$, and final normalization $\mathrm{norm}(\cdot)$, we use each probe $p$ as a raw continuation prompt, without a chat template, and record the last-position hidden state $h_\ell(p)$ at depths
$\ell \in \{L/4, L/2, 3L/4, L{-}2\}$.
We project each state through the model's output head,
\[
z_\ell(p) = W_U\,\mathrm{norm}(h_\ell(p)),
\]
following the logit-lens formulation \citep{nostalgebraist2020logitlens,belrose2023tuned}. For all 32 models, we also record the top-100 tokens from the final output logits.

We remove \emph{glitch tokens} whose unembedding norm is above $5\times$ or below $0.2\times$ the vocabulary median, decode the remaining top-100 token ids, strip surrounding whitespace, and discard empty strings. The fingerprint of a model is the mapping from each probe to its ranked list of decoded token strings. Extraction requires one forward pass per probe, with no sampling or gradients.

We retain the top 100 tokens to support the $k$ ablation, but use $k{=}20$ by default. This choice balances specificity against instability near the cutoff. Results vary by less than 0.04 in lineage group AUC over $k \in \{10,20,50,100\}$ (Appendix~C). We use unweighted token sets and leave rank-weighted variants to future work.

\subsection{Probe Suite}

The probe suite contains 250 prompts across 13 domains, including common and long-tail geography, science, history, code, mathematics, multi-step recall, culture, Chinese-language knowledge, multilingual prompts in nine languages, and contemporary facts. Per-domain counts and examples appear in Appendix~B, and the full suite is included in the released code.

The domain taxonomy and probe set were fixed before computing any cross-model similarity and were not modified in response to the results. Common-knowledge probes provide a shared baseline, while long-tail and language-specific probes contribute more strongly to discrimination. Domain-level aggregation is used to identify which content areas contribute to similarity between a given pair of models.

\subsection{Similarity, Null, and Statistics}
\label{sec:null}

For models $a$ and $b$ at depth $\ell$, we define similarity as the mean Jaccard overlap of their top-$k$ decoded-string sets across probes:
\[
S_\ell(a,b)
=
\frac{1}{|P|}
\sum_{p \in P}
\frac{|F_{a,\ell}(p)\cap F_{b,\ell}(p)|}
{|F_{a,\ell}(p)\cup F_{b,\ell}(p)|}.
\]
When $L_a \neq L_b$, internal depths are matched by fractional position.

String-level comparison permits evaluation across different vocabularies, but shared tokenizers and globally frequent tokens can inflate similarity independently of probe content. We therefore estimate a pair-specific \emph{mismatched-probe null},
$\tilde S(a,b)$, by pairing model $a$'s response to probe $i$ with model $b$'s response to a different probe $j \neq i$, using 200 random mismatched pairs. We report both the raw score $S$ and the \emph{excess similarity}
\[
S(a,b)-\tilde S(a,b).
\]

At near-final depth, the null averages 0.089 for same-tokenizer pairs and 0.054 for cross-tokenizer pairs; at $3L/4$, the corresponding values are 0.068 and 0.012. This correction removes probe-independent lexical overlap, but not differences caused by alternative tokenization of semantically equivalent strings. Cross-tokenizer similarity may therefore remain underestimated.

Statistical significance is assessed at the model level. We shuffle lineage-group labels over the 32 models for $10^4$ permutations and recompute the within-group versus between-group similarity gap, preserving dependence among pairwise scores. Cohen's $d$ is reported only as a descriptive pair-level effect size. Robustness analyses include leave-one-lineage-group-out AUC, removal of the full Qwen or Pythia blocks, and a leave-one-model-out jackknife.

\subsection{Model Pool}

The calibration pool contains 32 open-weight models ranging from 0.6B to 32B parameters and covering nine developer families; the full list is provided in Appendix~A. For analysis, we define 11 \emph{lineage groups}, separating R1-Qwen, R1-Llama, and Qwen2.5-Math from their broader developer families because their documented training histories differ.

The pool includes Qwen3 \citep{qwen3}, Qwen2.5 \citep{qwen25}, Llama-3.1/3.2 \citep{llama3}, Mistral and Ministral \citep{jiang2023mistral}, Phi-3.5 and Phi-4 \citep{abdin2024phi3,abdin2024phi4}, InternLM2.5 \citep{cai2024internlm2}, Gemma-2 \citep{gemma2}, Pythia \citep{biderman2023pythia,gao2020pile}, and five DeepSeek-R1 distillations \citep{deepseek2025r1}. It contains both base and post-trained models. The documented base checkpoints Qwen2.5-32B and Llama-3.1-8B are included in both the calibration and lineage-retrieval analyses.

Each experiment uses an explicitly defined pool. The \emph{calibration pool} contains the 32 models used for all ladder and lineage group-level analyses. The \emph{ancestry pool} adds Qwen2.5-Math-1.5B and Qwen2.5-14B as candidate bases, yielding 33 candidates for each query. The Pythia-deduped models and four additional base models---GPT-2-XL \citep{radford2019gpt2}, OPT-6.7B \citep{zhang2022opt}, OLMo-2-7B \citep{olmo2}, and Qwen2.5-7B---are used only as robustness controls and do not enter calibration statistics.

A separate \emph{witness pool} contains GPT-NeoX-20B \citep{black2022gptneox}, Cerebras-GPT 1.3--13B \citep{dey2023cerebras}, Pythia-2.8B, and RWKV-4-Pile 3B/7B \citep{peng2023rwkv}. These models provide same-corpus comparisons across organizations, tokenizers, and architecture classes and are used only in the witness analysis (Section~\ref{sec:witness}).

Near-final depth ($L{-}2$) is the primary internal readout. Final output distributions are available for all 32 calibration models and are used in the depth-ablation analysis.

\section{Results}

\subsection{The Calibration Ladder}

\begin{figure}[!htbp]
\centering
\includegraphics[width=\columnwidth]{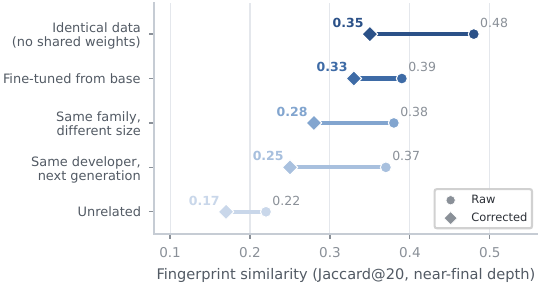}
\caption{Mean fingerprint similarity across documented relationship categories, shown for raw scores (circles) and excess scores after vocabulary correction (diamonds). The highest-similarity category contains no shared parameters.}
\label{fig:ladder}
\end{figure}

\begin{table}[!htbp]
\centering
\footnotesize
\setlength{\tabcolsep}{4pt}
\begin{tabular}{@{}lrcc@{}}
\toprule
 & $n$ & Raw & Excess\\
\midrule
\emph{Relationship categories} & & & \\
\quad Identical data, no shared weights & 3 & 0.48 & 0.35\\
\quad \emph{Replication: deduped suite}$^\dagger$ & 3 & 0.45 & 0.35\\
\quad Same corpus, diff.\ order (cross-size)$^\dagger$ & 6 & 0.46 & 0.35\\
\quad Same corpus, diff.\ order (size-matched)$^\dagger$ & 3 & 0.51 & 0.39\\
\quad Fine-tuned from documented base & 3 & 0.39 & 0.33\\
\quad Same lineage group, different size & 36 & 0.38 & 0.28\\
\quad Same developer, next generation & 30 & 0.37 & 0.25\\
\quad No documented relationship & 418 & 0.22 & 0.17\\
\addlinespace[2pt]
\emph{Lineage group structure} & & & \\
\quad ROC AUC (same vs.\ cross group) & & 0.859 & 0.847\\
\quad Permutation $p$ (model-level) & & $<10^{-4}$ & $<10^{-4}$\\
\quad Lineage retrieval (R1 distills) & 5 & \multicolumn{2}{c}{5/5 top-2; MRR 0.60}\\
\bottomrule
\end{tabular}
\caption{Results at near-final depth for the 32-model calibration pool. Probe-bootstrap 95\% CIs lie within $\pm 0.013$ of each category mean. The shared-base category contains the three distill$\leftrightarrow$base pairs for which the documented base is included in the pool: R1-Q-32B$\leftrightarrow$Q2.5-32B, R1-Q-7B$\leftrightarrow$Q2.5-Math-7B, and R1-L-8B$\leftrightarrow$L3.1-8B. The categories cover $490$ of the $\binom{32}{2}=496$ pairs. The remaining six pairs are among the four R1-Qwen distillations, which share a teacher but use different base models and are therefore excluded from the category analysis. $^\dagger$Control results from the Pythia-deduped suite, which is not included in the calibration pool; the corresponding $2{\times}2$ comparison isolates the effect of training order (Appendix~C).}
\label{tab:main}
\end{table}

Figure~\ref{fig:ladder} and Table~\ref{tab:main} summarize the main calibration result. Mean similarity generally increases with documented model relatedness for both raw and excess scores. The two same-developer categories are close (difference $+0.014$, 95\% CI $[+0.010,+0.019]$), so the results support three broader regimes: strongly related pairs, including identical-data and shared-base pairs (0.35 and 0.33 excess); same-developer pairs (0.25--0.28); and pairs with no documented relationship (0.17). Because the category distributions overlap, these values provide pool-level reference ranges rather than pair-level decision thresholds; individual retrieval results are evaluated using nearest-neighbor ranks and margins (\S\ref{sec:ancestry}).

The non-related category retains substantial similarity (0.22 raw; 0.17 excess), consistent with shared lexical and distributional structure across modern language models and motivating comparison against calibrated reference groups. The Pythia-deduped controls reproduce the identical-data result and isolate training order: cross-size similarity is nearly unchanged when the models use disjoint token streams and different orders (0.463 vs.\ 0.462). Within the resolution of this experiment, shared token order does not contribute detectably to the score (Appendix~C).

\begin{figure}[!h]
\centering
\includegraphics[width=\columnwidth]{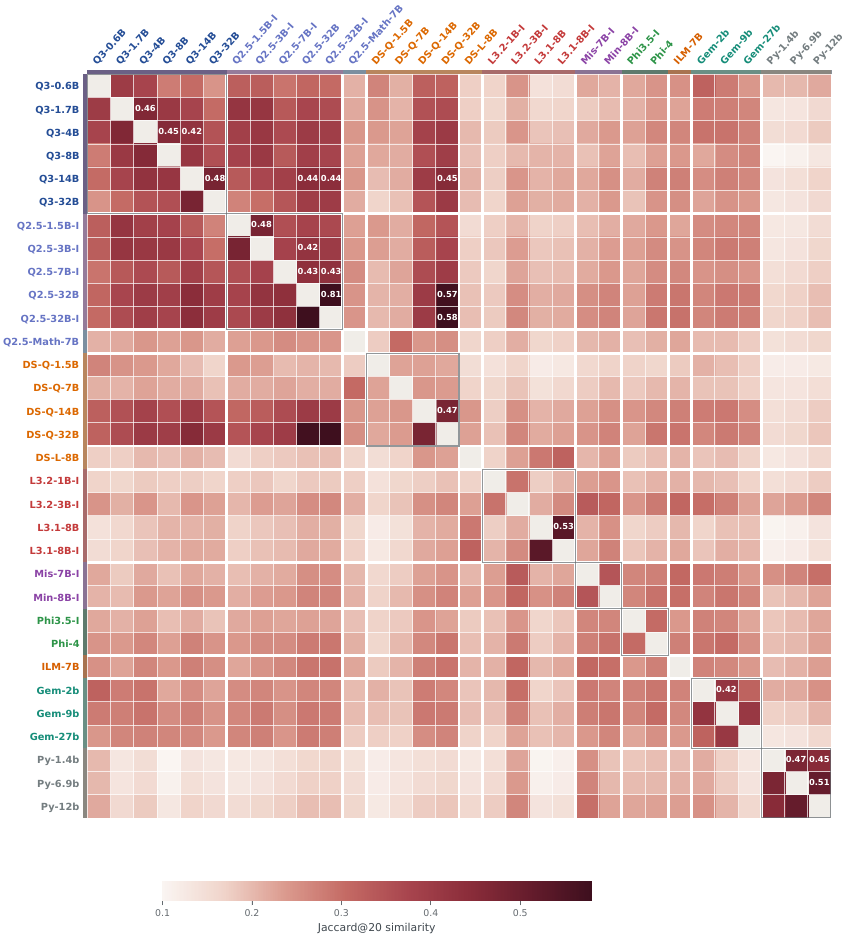}
\caption{Raw pairwise fingerprint similarity (Jaccard@20) at near-final depth for the 32 calibration models, ordered by lineage group. Within-lineage group blocks and several fine-tune$\leftrightarrow$base pairs are visible. The strongest non-identical pair is Qwen2.5-32B base$\leftrightarrow$Instruct (0.81); other high-similarity pairs include R1-Q-32B$\leftrightarrow$Qwen2.5 models (0.57--0.58) and the identical-data Pythia pairs (up to 0.51).}
\label{fig:matrix}
\end{figure}

\subsection{External Witnesses: The Same-Data Signal Across Organizations, Tokenizers, and Architectures}
\label{sec:witness}

The identical-data signal within Pythia could reflect shared tokenizer, training recipe, or token order rather than training data alone. We therefore evaluate three external witness groups that vary these factors (Table~\ref{tab:witness}).

GPT-NeoX-20B \citep{black2022gptneox}, trained on the Pile by the same organization as Pythia but in a separate project, achieves a mean excess similarity of 0.43 with Pythia at the output level, matching the Pythia-internal value. Cerebras-GPT \citep{dey2023cerebras} (1.3B--13B), trained on the Pile by a different organization with a different tokenizer and training recipe, reaches 0.35. This value is lower than the same-tokenizer witness results, consistent with reduced string overlap across tokenizers, but remains substantially above the reference value of 0.15 for pairs with no documented relationship. RWKV-4-Pile \citep{peng2023rwkv}, a recurrent non-Transformer model trained on the Pile, reaches 0.43 despite differing in tokenizer and architecture class. All witness comparisons use the output distribution for consistency.

\begin{table}[!htbp]
\centering
\footnotesize
\setlength{\tabcolsep}{3pt}
\begin{tabular}{@{}llcc@{}}
\toprule
Witness group & Main difference & $n$ & Excess\\
\midrule
Pythia internal & baseline & 3 & 0.43\\
NeoX-20B$\leftrightarrow$Pythia & project & 3 & 0.43\\
RWKV$\leftrightarrow$Pythia & architecture class & 6 & 0.43\\
Cerebras$\leftrightarrow$Pythia & tokenizer, organization, recipe & 12 & 0.35\\
\addlinespace[2pt]
No documented relationship & --- & 6 & 0.15\\
\bottomrule
\end{tabular}
\caption{Output-level excess similarity across same-corpus witness groups. The signal remains above the reference level across projects, organizations, tokenizers, and architecture classes. The lower Cerebras value is consistent with reduced decoded-string overlap across tokenizers.}
\label{tab:witness}
\end{table}

\begin{figure}[!htbp]
\centering
\includegraphics[width=\columnwidth]{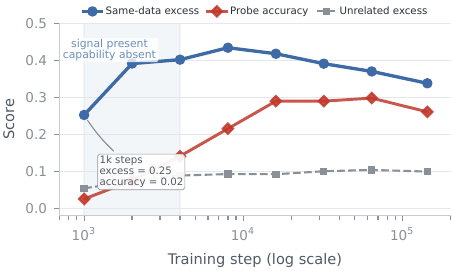}
\caption{Training trajectory on log-scaled steps. Excess similarity between Pythia-1.4B and Pythia-6.9B rises within the first 1\% of training, while probe accuracy remains near zero. Similarity to an external reference remains low.}
\label{fig:trajectory}
\end{figure}

We further examine when the signal emerges during training (Figure~\ref{fig:trajectory}). For 121 probes with unambiguous factual answers, we define \emph{probe accuracy} as the fraction for which a gold answer string appears among the model's top-20 decoded tokens. This measure uses the same forward pass as the fingerprint but evaluates factual answer recovery rather than cross-model overlap.

Across publicly released Pythia checkpoints, excess similarity between the 1.4B and 6.9B models reaches 0.25 at step 1000 (0.7\% of training), peaks at 0.43 by step 8000 (5.6\%), and declines to 0.34 at the final checkpoint. Over the same trajectory, similarity to Qwen2.5-7B remains between 0.05 and 0.10. At step 1000, probe accuracy is 0.02 while excess similarity is already 0.25. This temporal separation suggests that the early similarity signal is not explained solely by convergence in factual competence.

\subsection{Lineage Group Structure at the Model Level}

Within-group similarity exceeds cross-group similarity for every lineage group containing multiple models. A model-level permutation test yields $p<10^{-4}$ for both raw and excess scores, with observed gaps 10.2 and 9.5 standard deviations above the permutation mean, respectively. Pair-level Cohen's $d$ is 1.81 and is reported descriptively.

Using similarity to distinguish same-group from cross-group pairs gives an AUC of 0.859 for raw scores and 0.847 for excess scores. The result is not driven by a single model block: leave-one-lineage-group-out AUC ranges from 0.820 to 0.903; removing the full Qwen constellation gives 0.832; removing Pythia gives 0.820; and the leave-one-model-out jackknife ranges from 0.844 to 0.883 (Table~\ref{tab:robust}).

\begin{table}[t]
\centering
\footnotesize
\setlength{\tabcolsep}{3pt}
\renewcommand{\arraystretch}{1.05}

\begin{tabularx}{\columnwidth}{@{}
    >{\raggedright\arraybackslash}X
    >{\centering\arraybackslash}p{0.27\columnwidth}
    @{}}
\toprule
Configuration & Group AUC \\
\midrule
Full pool (32 models, raw) & 0.859 \\
Full pool (excess-corrected) & 0.847 \\
\addlinespace[2pt]
Leave-one-lineage-group-out (range) & 0.820--0.903 \\
Drop Qwen constellation & 0.832 \\
Drop Pythia & 0.820 \\
Leave-one-model-out jackknife (range) & 0.844--0.883 \\
\addlinespace[2pt]
$k=10,\ 20$ & 0.856,\ 0.859 \\
$k=50,\ 100$ & 0.839,\ 0.820 \\
\bottomrule
\end{tabularx}

\caption{Same-group discrimination across score correction, model-pool composition, resampling, and the choice of $k$ at near-final depth.}
\label{tab:robust}
\end{table}

\subsection{Same-Pool Baseline: Probability Overlap}
\label{sec:baseline}

Following the similarity used by PhyloLM \citep{yax2025phylolm}, we compute the Bhattacharyya overlap
$\sum_t \sqrt{P_a(t\,|\,p)P_b(t\,|\,p)}$
between output-token distributions and average it across probes. This baseline is evaluated on the 26-model subpool for which full probability vectors were retained.

On this subpool, probability overlap achieves a same-group AUC of 0.765, compared with 0.942 for the top-$k$ set fingerprint at the same output depth. In the single lineage-retrieval case whose documented base is available in this subpool, both methods rank the base first, with top-1 margins of $1.05\times$ and $1.09\times$, respectively. These results indicate that top-$k$ token identity is more discriminative on this pool, although the lineage comparison is limited to one query.

\subsection{Lineage Retrieval and Neighborhood Ambiguity}
\label{sec:ancestry}

For each of the five R1 distillations with an available documented base, we rank the remaining 33 models by fingerprint similarity (Table~\ref{tab:baseid}). The documented base appears within the top two in all five cases, with mean rank 1.8, MRR 0.60, and Hits@2 of 1.00. It ranks first in one case, R1-Q-7B$\to$Qwen2.5-Math-7B, with a margin of $1.27\times$. In the other four cases, the top-ranked candidate is a same-developer relative.

For R1-Q-32B and R1-L-8B, the Instruct variant associated with the same base checkpoint ranks above the base. For R1-Q-14B, another R1 distillation initialized from a different Qwen base ranks first. For R1-Q-1.5B, Qwen3-0.6B ranks first. These cases show that the fingerprint often identifies a local lineage neighborhood, but does not reliably distinguish the exact parent from closely related alternatives using output behavior alone.

The 7B query provides a more specific retrieval case. Its documented base is Qwen2.5-Math-7B rather than the general-purpose Qwen2.5-7B. A metadata baseline based on tokenizer lineage group and parameter count selects the wrong model, whereas the fingerprint ranks the math-specialized base first. This result suggests that behavioral similarity can distinguish some specialized and general-purpose variants that coarse metadata cannot.

\textbf{Candidate-removal analysis.}
We consider two candidate-removal conditions. In \emph{open-a}, the documented base is removed; the top-ranked candidate remains a same-developer relative in all five cases. In \emph{open-b}, the base, co-distilled siblings, and same-base Instruct variants are removed; four of five top-1 margins fall below $1.01\times$, while the remaining margin is $1.09\times$. These results suggest that small nearest-neighbor margins may indicate the absence of a close candidate, but a calibrated abstention rule would require an independent validation set.

\begin{table}[!htbp]
\centering
\scriptsize
\setlength{\tabcolsep}{2pt}
\renewcommand{\arraystretch}{1.08}

\begin{tabularx}{\columnwidth}{@{}
    >{\raggedright\arraybackslash}p{0.14\columnwidth}
    >{\raggedright\arraybackslash}p{0.25\columnwidth}
    >{\raggedright\arraybackslash}p{0.25\columnwidth}
    >{\centering\arraybackslash}p{0.09\columnwidth}
    >{\raggedright\arraybackslash}X
    @{}}
\toprule
Query & Top-1 & Top-2 & \shortstack{Base\\rank} & Relation \\
\midrule
R1-Q-32B
& Q2.5-32B-Inst (.58)
& \textbf{Q2.5-32B} (.57)
& 2
& same-base \\

R1-L-8B
& L3.1-8B-Inst (.32)
& \textbf{L3.1-8B} (.29)
& 2
& same-base \\

R1-Q-7B
& \textbf{Q2.5-Math-7B} (.31)
& R1-Q-14B (.24)
& 1
& exact \\

R1-Q-14B
& R1-Q-32B (.47)
& \textbf{Q2.5-14B} (.46)
& 2
& co-distill \\

R1-Q-1.5B
& Q3-0.6B (.27)
& \textbf{Q2.5-M-1.5B} (.26)
& 2
& next-gen \\
\bottomrule
\end{tabularx}

\caption{Lineage retrieval over 33 candidates at near-final depth. Bold denotes the documented base. The base ranks within the top two for all five queries, but ranks first in only one case. Relations describe the top-ranked candidate relative to the documented base.}
\label{tab:baseid}
\end{table}

\subsection{Depth Ablation and Operational Conditions}
\label{sec:depth}

\begin{figure}[!htbp]
\centering
\includegraphics[width=\columnwidth]{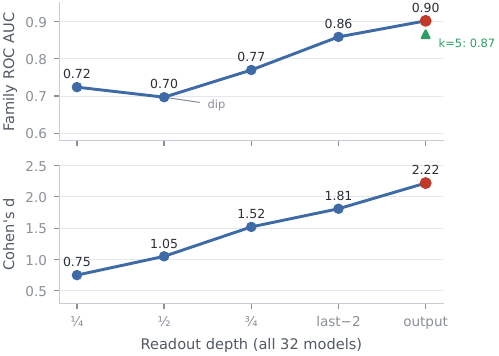}
\caption{Same-group discrimination across readout depths for all 32 calibration models. Performance is highest at the final output distribution; output-level AUC remains 0.87 when restricted to the top 5 tokens.}
\label{fig:depth}
\end{figure}

Figure~\ref{fig:depth} evaluates four internal readouts and the final output distribution across all 32 models. Same-group AUC is 0.72 at quarter depth, 0.70 at half depth, 0.77 at three-quarter depth, 0.86 at $L{-}2$, and 0.90 at the output. Pair-level Cohen's $d$ follows the same overall pattern and reaches 2.22 at the output.

The final output distribution is therefore the strongest readout in this evaluation. We nevertheless use $L{-}2$ for the primary internal-state analyses because fractional depths are not necessarily aligned across architectures and scales: the depth at which intermediate predictions approach the final distribution varies across model configurations (Appendix~C).

\textbf{Truncation and temperature.}
At the output level, restricting the fingerprint to the top $k{=}5$ tokens retains a same-group AUC of 0.87. Because the fingerprint depends only on token ranking, it is invariant to strictly monotone transformations of the logits, including temperature scaling applied before top-$k$ selection (Proposition~1, Appendix~\ref{app:formal}).

Domain-level aggregation provides a profile of which probe categories contribute to pairwise similarity. For example, Qwen$\leftrightarrow$Qwen similarity exceeds Qwen$\leftrightarrow$Llama similarity by $6.2\times$ on Chinese-language probes, whereas the difference is smaller on code probes (Appendix~C).

\section{Robustness}

\textbf{Tokenizer stratification.}
In the public model ecosystem, same-lineage group pairs generally also share a tokenizer, so cross-tokenizer same-lineage group AUC cannot be estimated directly because no positive pairs exist. We therefore assess tokenizer effects through three complementary analyses. First, within the 16-model Qwen subpool, whose models share the same BPE vocabulary core, lineage-group discrimination remains above chance (AUC 0.73) when separating Qwen2.5, Qwen3, and R1-Qwen models. Second, subtracting the mismatched-probe null reduces the contribution of probe-independent lexical overlap while retaining an AUC of 0.847. Third, the external witness analysis provides same-corpus comparisons across tokenizers: Cerebras$\leftrightarrow$Pythia pairs achieve an excess similarity of 0.35, compared with 0.15 for pairs with no documented relationship (Table~\ref{tab:witness}). These results indicate that tokenizer sharing contributes to raw similarity but does not fully explain the observed structure.

\textbf{Shared-vocabulary null.}
Table~\ref{tab:main} reports both raw and excess scores. The correction lowers absolute similarity while preserving the main within-group structure and the lineage-retrieval results. The estimated null is strongly tokenizer-dependent, averaging approximately 0.03 for Llama pairs and 0.13 for Qwen2.5 and Pythia pairs. Raw scores are therefore not directly comparable across tokenizer groups. Excess scores improve comparability by removing probe-independent overlap, although they do not correct for semantically equivalent strings that are segmented differently across vocabularies (Appendix~C).

\textbf{Base-model controls.}
We test whether the results could arise from systematic differences between base and instruction-tuned models. First, after adding Qwen2.5-7B, its base$\leftrightarrow$Instruct similarity is 0.76, close to the Qwen2.5-32B base$\leftrightarrow$Instruct value of 0.81. By contrast, the cross-lineage group pair Qwen2.5-7B$\leftrightarrow$Llama-3.1-8B-Instruct scores 0.20, near the reference level for pairs with no documented relationship. This suggests that post-training preserves substantially more of the fingerprint than changing model lineage.

Second, three additional base models---GPT-2-XL, OPT-6.7B, and OLMo-2-7B---form 24 cross-suite base$\leftrightarrow$base pairs with mean similarity 0.23, close to the overall reference mean of 0.22 and well below the Pythia identical-data mean of 0.48. Thus, a generic response pattern associated with base models does not reproduce the identical-data signal. The highest control similarities are GPT-2$\leftrightarrow$OPT (0.40) and OPT$\leftrightarrow$Pythia (0.34--0.37), consistent with their documented corpus relationships \citep{zhang2022opt,gao2020pile}; Qwen$\leftrightarrow$Pythia base pairs score 0.14--0.18.

\textbf{Stability.}
The main same-group discrimination result remains stable under leave-one-lineage-group-out evaluation, removal of the full Qwen or Pythia blocks, and a leave-one-model-out jackknife (Table~\ref{tab:robust}); each block-removal permutation test gives $p<10^{-3}$. Removing any single probe domain yields AUC values between 0.858 and 0.870, and the documented base remains within the top two for all five lineage queries. AUC changes by less than 0.04 across $k \in \{10,20,50,100\}$. Independent fp32 and bf16 extraction pipelines agree at Jaccard 0.98 (Appendix~C).

\textbf{Quantization.}
For DeepSeek-R1-Distill-Qwen-32B, the int8 fingerprint has Jaccard similarity 0.92 with the bf16 fingerprint, while int4 yields 0.85. For Qwen2.5-32B-Instruct, int4 yields 0.82. These values are comparable to or above the largest cross-model similarity in the calibration pool (0.81), although the Qwen2.5 int4 margin is only 0.01. The results indicate strong stability under int8 quantization and weaker, model-dependent stability under int4.

\textbf{Capability-convergence control.}
We fit the following regression:
\begin{equation}
\begin{aligned}
S^{\mathrm{excess}}_{ab}
={}& \beta_0
+ \beta_1\,\mathbb{I}[g_a=g_b] \\
&+ \beta_2\,\Delta C_{ab}
+ \beta_3\,\lvert \Delta \log P_{ab} \rvert
+ \varepsilon_{ab}.
\end{aligned}
\end{equation}
Using two-way cluster-robust standard errors at the model level, the same-group coefficient remains positive after controlling for capability and parameter scale
($\beta_1=0.103$, $z=6.46$).
Capability gap is also associated with similarity
($\beta_2=-0.24$, $z=-4.86$),
but capability gap alone yields an AUC of 0.49 for same-group discrimination.
These results indicate that measured capability similarity does not explain the observed lineage-group structure.

\section{Discussion and Limitations}

\textbf{Summary of evidence.}
The calibration and witness experiments support three conclusions. First, a strong fingerprint signal exists without weight sharing and persists across organizations, tokenizer families, and architecture classes (Table~\ref{tab:witness}). Second, the signal appears within the first 1\% of training, before measurable factual competence, and remains associated with lineage after controlling for capability and parameter scale (Figure~\ref{fig:trajectory}). Third, nearest-neighbor retrieval places all five R1 distillations within a small same-developer lineage neighborhood, including cases where coarse metadata does not identify the appropriate specialized base. Together, these findings indicate that token-space similarity captures persistent structure related to model training history rather than capability alone.

\textbf{Cross-tokenizer comparison.}
Decoded-string Jaccard enables comparison across vocabularies but can underestimate similarity when equivalent continuations are segmented differently. The witness results provide an empirical estimate of this effect: the same-data excess score is 0.43 for same-tokenizer comparisons and 0.35 for the cross-tokenizer Cerebras$\leftrightarrow$Pythia comparisons, while remaining well above the 0.15 reference level for pairs with no documented relationship.

\textbf{Behavioral interpretation.}
A scalar output-space similarity score does not by itself identify the mechanism underlying a relationship. Independent models trained on overlapping corpora may attain similarities comparable to weight-inheriting fine-tunes. We therefore interpret the fingerprint through the surrounding structure---candidate rankings, nearest-neighbor margins, domain profiles, and calibrated reference groups---rather than through an isolated pairwise score. In this role, the method is best suited to narrowing a candidate set and identifying which model relationships warrant further analysis.

\textbf{Scope.}
The lineage-retrieval evaluation covers five documented R1 distillations, while the same-data evidence spans multiple model suites, organizations, tokenizers, and architectures. Broader evaluations with additional lineage families and controlled training designs would further refine the calibration and its transfer across model ecosystems.

\section{Ethics and Responsible Use}

Model-provenance analysis may affect legal, commercial, or reputational decisions. Fingerprint scores should therefore be interpreted together with candidate rankings, margins, domain-level evidence, and independent provenance information where available. The method is intended to support model comparison and hypothesis generation rather than serve as standalone proof of a specific training history. We release the probes, code, and model fingerprints to enable independent verification and extension.

\section{Conclusion}

Top-$k$ token projections on a fixed probe set provide a lightweight fingerprint for comparing language models in output space. After calibration against documented relationships, the resulting similarity structure distinguishes broad levels of model relatedness, retrieves the documented base within the top two candidates for all five R1 distillations, and identifies specialized bases that coarse metadata can miss. The signal is strongest near the output distribution, remains effective with top-5 token access, and is stable under int8 and int4 quantization. It also persists across organizations, tokenizer families, and architecture classes, and emerges early in training. These results position token-space fingerprints as a practical tool for lineage retrieval and model-provenance analysis.

\bibliography{main}

\clearpage
\appendix

\section{Model Pool}
\label{app:pool}

The pool was constructed backward from the evidence requirements: every rung of the relatedness ladder needs pairs whose relationship is \emph{documented}, not inferred. Pythia supplies the identical-data rung (public training order, no weight sharing across sizes); the R1 distillations supply the shared-base rung (bases named in the release report); Qwen2.5$\to$Qwen3 supplies successive generations under one developer; and the remaining families supply same-lineage group and unrelated pairs. Sizes span 0.6B--32B so that scale is represented within families rather than confounded across them.

For analysis we track 11 \emph{lineage groups} rather than 9 developer families: R1-Qwen, R1-Llama, and Qwen2.5-Math are split off because their documented ancestry differs from their developers' main lines---R1 distills descend from Qwen/Llama bases but were post-trained by DeepSeek, and the Math models were continued-pretrained on a specialized corpus. Treating these as members of their surface families would contaminate both the lineage group statistics (a distill is not a sibling) and the ancestry experiment (the query would share a label with its own answer).

Each model carries exactly one role (table below). \emph{Calibration} models generate every lineage group and ladder statistic; \emph{ancestry candidates} are documented bases added only so the ancestry queries have their answers available; \emph{witness}, \emph{base-control}, \emph{dedup-control}, and \emph{quantization-control} models never enter calibration statistics---this separation is what keeps the headline numbers stable as evidence accumulates. All models are evaluated with identical raw continuation prompts and no chat template; the single exception is the teacher-probe experiment (Appendix~C), where the API constraint forces chat formatting and all local comparisons are re-extracted to match.

\input{appendix_models}

\section{Probe Suite}
\label{app:probes}

The suite balances two opposing requirements. \emph{Floor} probes (common geography, basic science, canonical history) are answered identically by every competent model; they contribute little discrimination but anchor the score---a model that fails them is out of distribution for the method, and their stability separates rank noise from real disagreement. \emph{Discriminative} probes (niche capitals, long-tail facts, Chinese-language knowledge, contemporary events) are where training distributions actually differ; ablations confirm these carry the lineage group signal (removing common-knowledge domains barely moves AUC, while the overall structure degrades if all long-tail domains are removed together). Code probes use raw multi-line continuation contexts rather than natural-language descriptions, so they probe code-corpus statistics rather than instruction-following. Multilingual probes (9 languages) and the 20 Chinese probes give the domain profile its resolution: they are the axis along which Qwen and Llama separate most sharply ($6.2\times$; Appendix~C).

Two disciplines govern the suite. First, it was authored from the domain taxonomy \emph{before any cross-model similarity was computed} and has not been modified since---no probe was added, removed, or reworded in response to results. Second, the 15 contemporary probes (e.g., current office-holders) date the fingerprint to a knowledge-cutoff window; comparisons across models with very different cutoffs should weight the leave-one-domain-out results (lineage group AUC 0.858--0.870 without any single domain, including this one), which show no conclusion depends on the temporal slice.

A 121-probe subset with unambiguous short factual answers (e.g., ``The capital of France is''~$\to$ Paris) doubles as the \emph{gold key} for the capability score used in the trajectory analysis and confound regression (\S\ref{sec:witness}): capability is the fraction of gold-key probes whose answer string appears in the model's top-20 tokens. The key was authored from probe text alone, blind to model outputs.

\input{appendix_probes}

The probe list follows; newline characters in code probes are rendered as \textbackslash n, and three probes in scripts without font support here (two Arabic, one Russian) are shown as placeholders---the complete list ships with the code release.
{\sloppy\raggedright
\input{appendix_probelist}
}

\section{Extended Robustness Results}
\label{app:robust}

\begin{figure*}[t]
\centering
\includegraphics[width=0.88\textwidth]{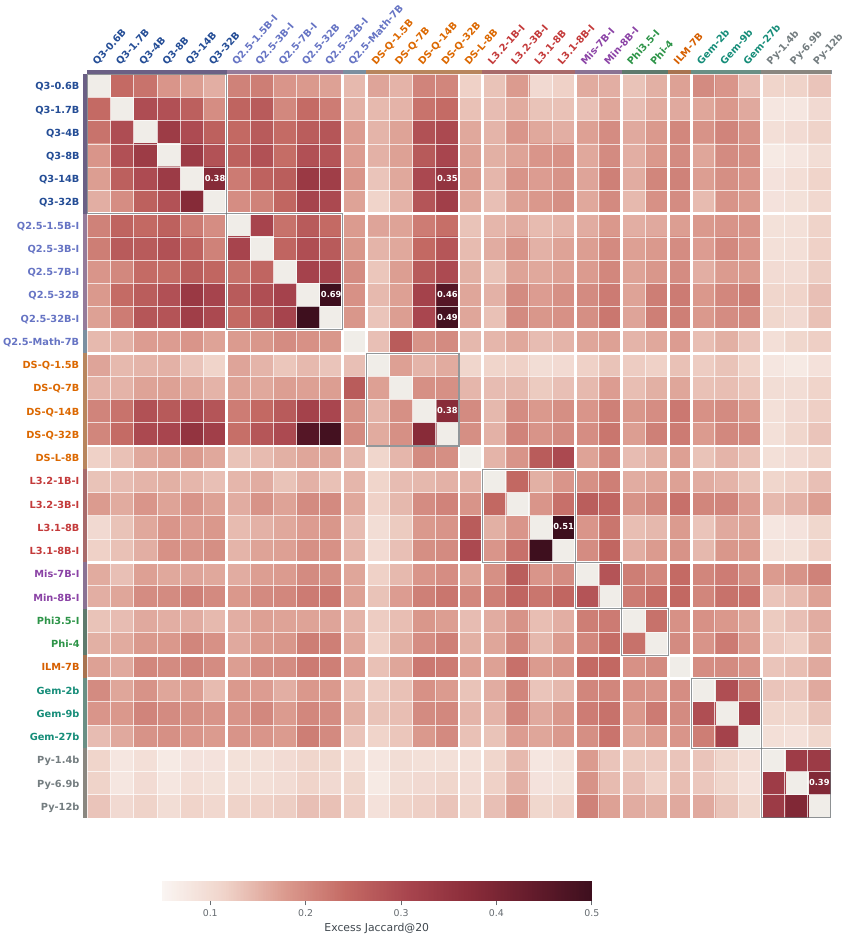}
\caption{Excess-corrected similarity matrix (matched minus mismatched-probe null, near-final depth). Lineage group blocks, the fine-tune pairs, and the Pythia island persist after the shared-vocabulary correction; what the correction changes is the cross-lineage-group comparability of the values (Table~\ref{tab:famnull}).}
\label{fig:matrix_excess}
\end{figure*}

\subsection*{What the Correction Does to the Matrix}

Comparing the raw matrix (Figure~\ref{fig:matrix}) with the corrected one (Figure~\ref{fig:matrix_excess}) shows that the correction is a per-pair \emph{recentering}, not a compression: every lineage group block, the fine-tune diagonal pairs, and the Pythia island survive, and every multi-member lineage group's intra-lineage-group mean stays above the corrected unrelated baseline of 0.166 (Table~\ref{tab:famnull}). What changes is comparability. The null component varies by a factor of five across families---0.027 for Llama pairs against 0.128 for Qwen2.5 and Pythia pairs---because it is driven by how much probability-mass-agnostic string overlap two vocabularies produce, which depends on vocabulary size, shared BPE merges, and each lineage group's propensity for generic high-frequency continuations. Raw values therefore systematically overstate the coherence of large-shared-vocabulary families relative to families with small or idiosyncratic token inventories. The correction costs a small amount of global discrimination (AUC 0.859$\to$0.847, because cross-tokenizer unrelated pairs are easy negatives in raw scores) and buys interpretability: corrected values mean the same thing on both sides of a tokenizer boundary, which is the property the ladder and the witness comparisons rely on.

\begin{table}[!htbp]
\centering
\footnotesize
\setlength{\tabcolsep}{5pt}
\begin{tabular}{@{}lrccc@{}}
\toprule
lineage group & $n$ & Raw & Null & Excess\\
\midrule
Pythia & 3 & 0.476 & 0.127 & 0.349\\
Qwen2.5 & 10 & 0.446 & 0.128 & 0.318\\
Gemma-2 & 3 & 0.383 & 0.110 & 0.274\\
Qwen3 & 15 & 0.374 & 0.114 & 0.260\\
Mistral & 1 & 0.343 & 0.062 & 0.280\\
Phi & 1 & 0.303 & 0.071 & 0.232\\
DS-Qwen & 6 & 0.271 & 0.065 & 0.206\\
Llama & 6 & 0.280 & 0.027 & 0.253\\
\bottomrule
\end{tabular}
\caption{Per-lineage-group decomposition of intra-lineage-group similarity into null and excess components (near-final depth; $n$ = intra-lineage-group pairs). The null spans $0.027$--$0.128$ across families, so raw values are not comparable across tokenizer groups; excess values are. All families remain above the corrected unrelated baseline (0.166).}
\label{tab:famnull}
\end{table}

\textbf{Training-order $2{\times}2$ (Pythia-deduped suite).} The deduped suite completes a $2{\times}2$ over token order and size matching: cross-size pairs score 0.463 with identical token streams and 0.462 with disjoint streams and orders---a null order effect isolated with size held constant---while matched-size cross-suite pairs reach 0.51, an elevation attributable to scale matching (consistent with the convergence-depth effect below) rather than shared order. The order conclusion is immune to possible seed sharing because its cells compare across sizes, where initializations necessarily differ.

\begin{figure}[!htbp]
\centering
\includegraphics[width=\columnwidth]{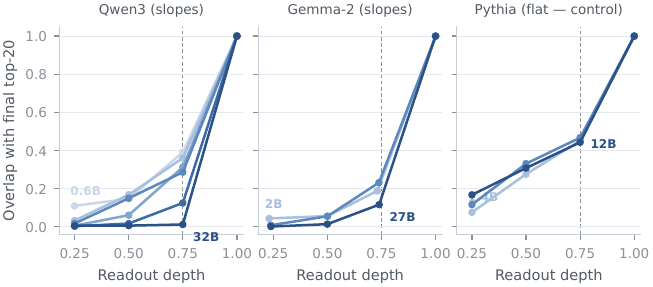}
\caption{Self-overlap between each depth's top-20 and the model's own final top-20. Within Qwen3 and Gemma-2, convergence shifts later with scale; within Pythia (identical data, order, and recipe across sizes) it does not shift, implicating training configuration rather than parameter count alone.}
\label{fig:cryst}
\end{figure}

\textbf{Prediction-convergence depth.} Comparing each model's intermediate top-20 with its own final top-20 measures where along the depth axis the eventual prediction takes shape (Figure~\ref{fig:cryst}). In Qwen3, self-overlap at $3L/4$ falls with scale (0.39 at 0.6B to 0.01 at 32B); Gemma-2 replicates the trend (0.23 at 9B, 0.12 at 27B). The Pythia suite provides the control: across 1.4B, 6.9B, and 12B trained on identical data in identical order, self-overlap at $3L/4$ is flat (0.45, 0.47, 0.44). Whatever shifts convergence later co-varies with how larger models are trained, not merely with their size---and this misalignment of fractional depth across scales is why all cross-model results in the paper use near-final depth.

\textbf{Domain profiles.} On Chinese-language probes, Qwen$\leftrightarrow$Qwen similarity exceeds Qwen$\leftrightarrow$Llama by $6.2\times$ (probe-bootstrap 95\% CI $4.7$--$8.4\times$, at $3L/4$ over four Qwen and two Llama models); on code probes the same comparison largely closes (0.14 vs.\ 0.08).

\begin{figure}[!htbp]
\centering
\includegraphics[width=\columnwidth]{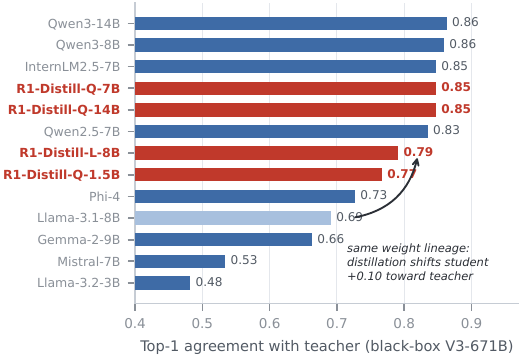}
\caption{Black-box teacher probe: top-1 agreement of 13 chat-aligned local models with the sampled first-token behavior of DeepSeek-V3 (671B). Red: R1-distilled students. Group-level agreement is dominated by a shared-lab-data cluster, but the backbone-controlled pair (arrow) shows distillation displacing the Llama-based student $+0.10$ toward the teacher under zero weight sharing.}
\label{fig:teacher}
\end{figure}

\textbf{Corrected ladder.} Excess-corrected category means: identical data 0.349, fine-tuned from base ($n{=}3$) 0.332, same lineage group 0.276, successive generation 0.248, unrelated 0.166. The mismatched-probe null averages 0.090 for same-tokenizer pairs and 0.052 across tokenizers at near-final depth.

\textbf{Corrected ancestry identification.} Under excess similarity the outcome matches the raw analysis: the documented base is top-1 for one of five R1 distillations and top-2 for all five. The lineage-neighborhood ambiguity persists under correction: in the four non-top-1 cases, the displacing candidate remains the same (same-base sibling, co-distillation, or next-generation model), with similar margins.

\begin{table*}[!htbp]
\centering
\scriptsize
\setlength{\tabcolsep}{8pt}
\begin{tabular}{@{}l@{\hspace{10pt}}lllll@{}}
\toprule
Query & Rank 1 & Rank 2 & Rank 3 & Rank 4 & Rank 5\\
\midrule
R1-Qwen-1.5B & Q3-0.6B .27 & \textbf{Q2.5-M-1.5B} .26 & Q3-1.7B .25 & Q2.5-1.5B .24 & Q3-4B .24\\
R1-Qwen-7B & \textbf{Q2.5-M-7B} .31 & R1-Qwen-14B .24 & R1-Qwen-32B .24 & R1-Qwen-1.5B .23 & Q3-4B .23\\
R1-Qwen-14B & R1-Qwen-32B .47 & \textbf{Q2.5-14B} .46 & Q2.5-32B-I .40 & Q3-14B .40 & Q3-4B .38\\
R1-Qwen-32B & Q2.5-32B-I .58 & \textbf{Q2.5-32B} .57 & R1-Qwen-14B .47 & Q2.5-14B .45 & Q3-14B .45\\
R1-Llama-8B & L3.1-8B-I .32 & \textbf{L3.1-8B} .29 & R1-Qwen-14B .24 & L3.2-3B .23 & Min-8B .23\\
\bottomrule
\end{tabular}
\caption{Full top-5 candidate rankings for lineage retrieval over 33 candidates (raw, near-final depth). Bold: documented base. When the base is not rank 1, the displacing candidate is a same-developer relative: same-base Instruct sibling (marked ``-I''), co-distillation, or next-generation model. The sole cross-developer entrant is Ministral-8B in the R1-Llama-8B list (rank 5, at the unrelated-baseline level 0.23).}
\label{tab:top5}
\end{table*}

\textbf{Model-level significance.} The permutation test shuffles lineage group labels over models, preserving all pairwise dependence under the null: raw $p<10^{-4}$ (observed gap 10.2$\sigma$ above the null), corrected $p<10^{-4}$ (9.5$\sigma$). Dropping the Qwen constellation or Pythia leaves $p<10^{-3}$ at $5{,}000$ permutations.

\textbf{Glitch filtering.} The unembedding-norm filter removes 0\% of the vocabulary for most models and at most 1.6\% (Qwen3-14B). Results are insensitive to the threshold because filtered tokens essentially never enter a top-100 list on our probes.

\textbf{Teacher-side probe (black-box).} DeepSeek-V3 was fingerprinted through its serving API by sampling 24 first tokens per probe at temperature 1 (no log-probabilities are exposed for this model lineage group). Because the API applies a chat template, all 13 local comparison models were re-extracted under their own chat templates, with thinking-mode blocks explicitly closed so the measured token is the answer token for every model. Top-1 agreement with the teacher: Qwen3-14B/8B 0.86, InternLM 0.85, R1-Distill-Qwen-7B/14B 0.85, Qwen2.5-7B 0.83, R1-Distill-Llama-8B 0.79, R1-Distill-Qwen-1.5B 0.77, Phi-4 0.73, Llama-3.1-8B 0.69, Gemma-9B 0.66, Mistral-7B 0.53, Llama-3.2-3B 0.48. The distilled-student group does not separate from the Chinese-lab data cluster as a whole (model-level permutation $p=0.13$, $n=13$); the backbone-controlled contrast (R1-Distill-Llama-8B vs.\ its base Llama-3.1-8B) shows the $+0.10$ displacement reported in the Discussion.

\textbf{Prediction-convergence values.} Self-overlap at $3L/4$ with the final top-20: Qwen3 0.6B/1.7B/4B/8B/14B/32B $=$ 0.39/0.36/0.31/0.29/0.13/0.01; Gemma-2 2B/9B/27B $=$ 0.19/0.23/0.12; Pythia 1.4B/6.9B/12B $=$ 0.45/0.47/0.44.

\section{Formal Properties of the Score}
\label{app:formal}

Fix a probe set $\mathcal{P}$ and $k \in \mathbb{N}$. For a model $m$ with vocabulary $V_m$, let $z_m(p) \in \mathbb{R}^{|V_m|}$ denote its logits (at a given readout) on probe $p$, and let $F_m(p) \subset \Sigma^*$ be the set of decoded, whitespace-stripped strings of the top-$k$ entries of $z_m(p)$. The score is
\[
S(a,b) \;=\; \frac{1}{|\mathcal{P}|}\sum_{p \in \mathcal{P}} J\!\big(F_a(p),\, F_b(p)\big),
\qquad J(A,B) = \frac{|A \cap B|}{|A \cup B|}.
\]

\textbf{Proposition 1 (invariance under monotone score transformations).}
\emph{Let $g:\mathbb{R}\to\mathbb{R}$ be strictly increasing and let $z'_m(p) = g\big(z_m(p)\big)$ entrywise. Then $F'_m(p) = F_m(p)$ for every $p$, and hence $S$ is unchanged.}

\emph{Proof.} A strictly increasing $g$ preserves the ordering of logit entries, so the top-$k$ index set---and therefore the decoded string set---is identical. \hfill$\square$

Temperature scaling ($z/\tau$, $\tau>0$), additive shifts, and softmax renormalization are all monotone, so the fingerprint is exactly invariant to them. This holds for any access mode that returns token \emph{identities} in rank order (full logits, ranked log-probabilities, top-$k$ lists); it does not extend to sampled tokens, whose distribution does change with temperature.

\textbf{The mismatched-probe null.}
The null $\tilde S(a,b)$ is the same statistic computed on probe pairs $(p_i, p_j)$, $i \ne j$, drawn uniformly at random. Its interpretation rests on one assumption: under the hypothesis that a model's top-$k$ set does not depend on probe content, matched and mismatched pairings are exchangeable, so $\mathbb{E}[S] = \mathbb{E}[\tilde S]$ and the excess $S - \tilde S$ estimates the component of similarity that is \emph{locked to probe content}. The null is far from the value implied by uniform random sets: two independent uniform $k$-subsets of a vocabulary of size $|V|$ have $\mathbb{E}|A\cap B| = k^2/|V|$, giving expected Jaccard $\approx 6.7\times10^{-5}$ for $k{=}20$, $|V|{=}150\text{k}$ (and $\approx 2\times10^{-4}$ for $|V|{=}50\text{k}$)---three orders of magnitude below the observed nulls of $0.054$--$0.089$. The null is therefore driven almost entirely by the frequency concentration of natural-language token distributions, not by vocabulary size, which is why it must be estimated empirically per pair rather than derived analytically, and why it varies by a factor of four across families (Table~\ref{tab:famnull}).

\textbf{Proposition 2 (mechanism non-identifiability of behavioral scores).}
\emph{Let $\Phi(m) = \{P_m(\cdot \mid p)\}_{p \in \mathcal{P}}$ be the map from a model to its probe-conditional output distributions, and let $S$ be any score of the form $S(a,b) = f\big(\Phi(a), \Phi(b)\big)$. If two generative histories---(i) $b$ inherits weights from $a$; (ii) $a$ and $b$ are trained independently on overlapping corpora---can produce the same pair $\big(\Phi(a), \Phi(b)\big)$, then no such $S$ distinguishes them.}

\emph{Proof.} Immediate: $S$ factors through $\Phi$, and $\Phi$ does not encode parameter history. \hfill$\square$

The proposition is elementary, but it delimits what \emph{any} behavioral score---ours, PhyloLM's, or a future one---can certify: behavioral evidence identifies training-distribution overlap, not the mechanism that produced it. Our pool contains an empirical witness that the antecedent is realized in practice: GPT-2$\leftrightarrow$OPT (independent training, overlapping web corpora) scores 0.40, the same value as the weight-inheriting fine-tune rung. This is why the paper's inferential unit is never a single pairwise score but structure around it---nearest-neighbor margins against a candidate pool, domain profiles, and the calibrated ladder---and why mechanism claims additionally require weight-space evidence where weights are available.

\section{Reproducibility}
\label{app:repro}

All experiments run on a single 80GB GPU or smaller. Extraction is one forward pass per probe per model; models up to 14B ran locally in fp16/fp32; the 27--32B models and the two exact base checkpoints (Qwen2.5-32B, Llama-3.1-8B) ran in bf16 on rented A100 instances. All 32 models thus have both internal-depth fingerprints and output-level top-100 fingerprints; the distinction matters only for the Bhattacharyya baseline (\S\ref{sec:baseline}), which requires full probability vectors and is therefore restricted to the 26 models whose complete logit tensors were saved locally. A paired extraction of one model in fp32 and bf16 through two independent code paths agrees at Jaccard 0.98. Total extraction compute for the pool is under four GPU-hours (including the rental sessions). Probes, extraction and analysis code, per-model fingerprints, and all similarity matrices are released with the paper.
\end{document}

%% file: appendix_models.tex
\begin{table*}[!htbp]
\centering
\footnotesize
\setlength{\tabcolsep}{3pt}
\begin{tabular}{@{}llrrl@{}}
\toprule
Model & Family & $L$ & $|V|$ & Variant\\
\midrule
Cerebras-GPT-1.3B & Cerebras & 24 & 50k & base / witness pool\\
Cerebras-GPT-2.7B & Cerebras & 32 & 50k & base / witness pool\\
Cerebras-GPT-6.7B & Cerebras & 32 & 50k & base / witness pool\\
Cerebras-GPT-13B & Cerebras & 40 & 50k & base / witness pool\\
\addlinespace[2pt]
DeepSeek-R1-Distill-Llama-8B & DS-Llama & 32 & 128k & post-trained / calibration\\
\addlinespace[2pt]
DeepSeek-R1-Distill-Qwen-1.5B & DS-Qwen & 28 & 151k & post-trained / calibration\\
DeepSeek-R1-Distill-Qwen-7B & DS-Qwen & 28 & 152k & post-trained / calibration\\
DeepSeek-R1-Distill-Qwen-14B & DS-Qwen & 48 & 152k & post-trained / calibration\\
DeepSeek-R1-Distill-Qwen-32B-int4 & DS-Qwen & 64 & 152k & int4 / quant. control\\
DeepSeek-R1-Distill-Qwen-32B-int8 & DS-Qwen & 64 & 152k & int8 / quant. control\\
DeepSeek-R1-Distill-Qwen-32B & DS-Qwen & 64 & 152k & post-trained / calibration\\
\addlinespace[2pt]
gpt2-xl & GPT2 & 48 & 50k & base / base control\\
\addlinespace[2pt]
gemma-2-2b-it & Gemma & 26 & 256k & post-trained / calibration\\
gemma-2-9b-it & Gemma & 42 & 256k & post-trained / calibration\\
gemma-2-27b-it & Gemma & 46 & 256k & post-trained / calibration\\
\addlinespace[2pt]
internlm2\_5-7b-chat & InternLM & 32 & 92k & post-trained / calibration\\
\addlinespace[2pt]
Llama-3.2-1B-Instruct & Llama & 16 & 128k & post-trained / calibration\\
Llama-3.2-3B-Instruct & Llama & 28 & 128k & post-trained / calibration\\
Llama-3.1-8B-Instruct & Llama & 32 & 128k & post-trained / calibration\\
Llama-3.1-8B & Llama & 32 & 128k & base / calibration\\
\addlinespace[2pt]
Mistral-7B-Instruct-v0.3 & Mistral & 32 & 32k & post-trained / calibration\\
Ministral-8B-Instruct-2410 & Mistral & 36 & 131k & post-trained / calibration\\
\addlinespace[2pt]
gpt-neox-20b & NeoX & 44 & 50k & base / witness pool\\
\addlinespace[2pt]
OLMo-2-1124-7B & OLMo & 32 & 100k & base / base control\\
\addlinespace[2pt]
opt-6.7b & OPT & 32 & 50k & base / base control\\
\addlinespace[2pt]
Phi-3.5-mini-instruct & Phi & 32 & 32k & post-trained / calibration\\
phi-4 & Phi & 40 & 100k & post-trained / calibration\\
\addlinespace[2pt]
pythia-1.4b & Pythia & 24 & 50k & base / calibration\\
pythia-2.8b & Pythia & 32 & 50k & base / witness pool\\
pythia-6.9b & Pythia & 32 & 50k & base / calibration\\
pythia-12b & Pythia & 36 & 50k & base / calibration\\
\addlinespace[2pt]
pythia-1.4b-deduped & Pythia-dd & 24 & 50k & base / dedup control\\
pythia-6.9b-deduped & Pythia-dd & 32 & 50k & base / dedup control\\
pythia-12b-deduped & Pythia-dd & 36 & 50k & base / dedup control\\
\addlinespace[2pt]
Qwen2.5-Math-1.5B & Q25-Math & 28 & 151k & base (math) / ancestry cand.\\
Qwen2.5-Math-7B & Q25-Math & 28 & 152k & base (math) / calibration\\
\addlinespace[2pt]
Qwen2.5-1.5B-Instruct & Qwen2.5 & 28 & 151k & post-trained / calibration\\
Qwen2.5-3B-Instruct & Qwen2.5 & 36 & 151k & post-trained / calibration\\
Qwen2.5-7B-Instruct & Qwen2.5 & 28 & 152k & post-trained / calibration\\
Qwen2.5-7B & Qwen2.5 & 28 & 152k & base / base control\\
Qwen2.5-14B & Qwen2.5 & 48 & 152k & base / ancestry cand.\\
Qwen2.5-32B-Instruct-int4 & Qwen2.5 & 64 & 152k & int4 / quant. control\\
Qwen2.5-32B-Instruct & Qwen2.5 & 64 & 152k & post-trained / calibration\\
Qwen2.5-32B & Qwen2.5 & 64 & 152k & base / calibration\\
\addlinespace[2pt]
Qwen3-0.6B & Qwen3 & 28 & 151k & post-trained / calibration\\
Qwen3-1.7B & Qwen3 & 28 & 151k & post-trained / calibration\\
Qwen3-4B & Qwen3 & 36 & 151k & post-trained / calibration\\
Qwen3-8B & Qwen3 & 36 & 151k & post-trained / calibration\\
Qwen3-14B & Qwen3 & 40 & 151k & post-trained / calibration\\
Qwen3-32B & Qwen3 & 64 & 151k & post-trained / calibration\\
\addlinespace[2pt]
rwkv-4-3b-pile & RWKV & 32 & 50k & base / witness pool (output only)\\
rwkv-4-7b-pile & RWKV & 32 & 50k & base / witness pool (output only)\\
\bottomrule
\end{tabular}
\caption{All models with their roles. \emph{calibration}: the 32-model pool behind every family/ladder statistic; \emph{ancestry cand.}: documented-base additions used only as ancestry candidates; \emph{witness pool}: same-data witnesses from independent organizations (\S\ref{sec:witness}), excluded from calibration statistics; \emph{base control} / \emph{dedup control}: Robustness-section controls; \emph{quant.\ control}: quantized variants of calibration models, used only in the quantization-envelope analysis. RWKV witnesses are non-Transformer and contribute output-level fingerprints only. The black-box teacher DeepSeek-V3 (671B; Appendix~C) is accessed via its serving API and hosts no local checkpoint. $L$: layers; $|V|$: vocabulary size.}
\label{tab:pool}
\end{table*}

%% file: appendix_probes.tex
\begin{table*}[!htbp]
\centering
\footnotesize
\setlength{\tabcolsep}{3pt}
\begin{tabular}{@{}lrl@{}}
\toprule
Domain & $n$ & Example probe\\
\midrule
Geography (common) & 20 & \emph{The capital of France is}\\
Geography (niche) & 20 & \emph{The capital of Burkina Faso is}\\
Science (common) & 18 & \emph{The chemical formula for water is}\\
Science (niche) & 17 & \emph{The Chandrasekhar limit for white dwarf stars }\\
Code / Technical & 30 & \emph{def fibonacci(n):     if n \textless{}= 1:         retur}\\
History (common) & 13 & \emph{The first president of the United States was}\\
History (niche) & 12 & \emph{The Treaty of Westphalia was signed in}\\
Math / Reasoning & 25 & \emph{The square root of 144 is}\\
Culture / Language & 20 & \emph{The word 'Schadenfreude' means}\\
Chinese-specific & 20 & \begin{CJK}{UTF8}{gbsn}\emph{中国最长的河流是}\end{CJK}\\
Multilingual & 20 & \emph{La capitale de l'Allemagne est}\\
Temporal / Contemporary & 15 & \emph{The CEO of OpenAI is}\\
Long-tail / Obscure & 20 & \emph{The national animal of Scotland is}\\
\midrule
Total & 250 & \\
\bottomrule
\end{tabular}
\caption{Probe suite composition. Probes were authored from this domain taxonomy before any cross-model similarity was computed and never modified afterward; the complete list ships with the code release.}
\label{tab:probes}
\end{table*}

%% file: appendix_probelist.tex
{\footnotesize

\paragraph{Geography (common).}
\emph{The capital of France is} \;$\cdot$\; \emph{The capital of Japan is} \;$\cdot$\; \emph{The capital of Brazil is} \;$\cdot$\; \emph{The capital of Australia is} \;$\cdot$\; \emph{The capital of Egypt is} \;$\cdot$\; \emph{The capital of India is} \;$\cdot$\; \emph{The capital of Canada is} \;$\cdot$\; \emph{The capital of South Korea is} \;$\cdot$\; \emph{The capital of Mexico is} \;$\cdot$\; \emph{The capital of Germany is} \;$\cdot$\; \emph{The largest country by area is} \;$\cdot$\; \emph{The longest river in the world is} \;$\cdot$\; \emph{The highest mountain in the world is} \;$\cdot$\; \emph{The largest ocean on Earth is} \;$\cdot$\; \emph{The smallest continent by area is} \;$\cdot$\; \emph{The largest desert in the world is} \;$\cdot$\; \emph{The country with the largest population is} \;$\cdot$\; \emph{The deepest point in the ocean is called} \;$\cdot$\; \emph{The largest island in the world is} \;$\cdot$\; \emph{The longest wall ever built is} \;$\cdot$\; 
\paragraph{Geography (niche).}
\emph{The capital of Burkina Faso is} \;$\cdot$\; \emph{The capital of Bhutan is} \;$\cdot$\; \emph{The capital of Moldova is} \;$\cdot$\; \emph{The capital of Suriname is} \;$\cdot$\; \emph{The capital of Liechtenstein is} \;$\cdot$\; \emph{The second largest city in Kazakhstan is} \;$\cdot$\; \emph{The deepest lake in the world is located in} \;$\cdot$\; \emph{The smallest country in Africa by area is} \;$\cdot$\; \emph{The highest capital city in the world is} \;$\cdot$\; \emph{The longest river in Europe is} \;$\cdot$\; \emph{The largest landlocked country is} \;$\cdot$\; \emph{The country with the most time zones is} \;$\cdot$\; \emph{The driest inhabited continent is} \;$\cdot$\; \emph{The largest freshwater lake by surface area is} \;$\cdot$\; \emph{The only country that borders both the Atlantic and Indian oceans is} \;$\cdot$\; \emph{The strait separating Europe from Asia is called} \;$\cdot$\; \emph{The second highest mountain in the world is} \;$\cdot$\; \emph{The country with the most UNESCO World Heritage Sites is} \;$\cdot$\; \emph{The largest volcanic island is} \;$\cdot$\; \emph{The capital of the Maldives is} \;$\cdot$\; 
\paragraph{Science (common).}
\emph{The chemical formula for water is} \;$\cdot$\; \emph{The chemical symbol for gold is} \;$\cdot$\; \emph{The number of elements in the periodic table is approximately} \;$\cdot$\; \emph{The speed of light in meters per second is approximately} \;$\cdot$\; \emph{The force of gravity on Earth is approximately} \;$\cdot$\; \emph{The boiling point of water at sea level in Celsius is} \;$\cdot$\; \emph{The number of chromosomes in a human cell is} \;$\cdot$\; \emph{The powerhouse of the cell is called the} \;$\cdot$\; \emph{The molecule that carries genetic information is} \;$\cdot$\; \emph{The largest organ in the human body is} \;$\cdot$\; \emph{The closest star to Earth is} \;$\cdot$\; \emph{The number of planets in our solar system is} \;$\cdot$\; \emph{The largest planet in our solar system is} \;$\cdot$\; \emph{The unit of electrical resistance is the} \;$\cdot$\; \emph{The pH of pure water at room temperature is} \;$\cdot$\; \emph{The process by which plants convert sunlight to energy is} \;$\cdot$\; \emph{The speed of sound in air at room temperature is approximately} \;$\cdot$\; \emph{The distance from the Earth to the Sun is approximately} \;$\cdot$\; 
\paragraph{Science (niche).}
\emph{The Chandrasekhar limit for white dwarf stars is approximately} \;$\cdot$\; \emph{The enzyme that unwinds DNA during replication is called} \;$\cdot$\; \emph{The Krebs cycle produces a net total of} \;$\cdot$\; \emph{The oxidation state of manganese in potassium permanganate is} \;$\cdot$\; \emph{The fine-structure constant alpha is approximately} \;$\cdot$\; \emph{The neurotransmitter primarily responsible for reward is} \;$\cdot$\; \emph{The IUPAC name for aspirin is} \;$\cdot$\; \emph{The Schwarzschild radius of the Sun is approximately} \;$\cdot$\; \emph{The number of ATP molecules produced by oxidative phosphorylation is approximately} \;$\cdot$\; \emph{The electron configuration of chromium is} \;$\cdot$\; \emph{The de Broglie wavelength of an electron at 100 eV is approximately} \;$\cdot$\; \emph{The enzyme responsible for adding nucleotides during DNA replication is} \;$\cdot$\; \emph{The Hubble constant is approximately} \;$\cdot$\; \emph{The critical temperature of high-Tc superconductor YBCO is approximately} \;$\cdot$\; \emph{The protein that transports oxygen in blood is} \;$\cdot$\; \emph{The crystal field splitting in an octahedral complex is denoted} \;$\cdot$\; \emph{The Roche limit for a fluid satellite is approximately} \;$\cdot$\; 
\paragraph{Code / Technical.}
\emph{def fibonacci(n): \textbackslash{}n     if n \textless{}= 1: \textbackslash{}n         return} \;$\cdot$\; \emph{def binary\_search(arr, target): \textbackslash{}n     left, right = 0, len(arr) - 1 \textbackslash{}n     while} \;$\cdot$\; \emph{import torch \textbackslash{}n model = torch.nn.Linear(} \;$\cdot$\; \emph{with open('data.json', 'r') as f: \textbackslash{}n     data =} \;$\cdot$\; \emph{class Node: \textbackslash{}n     def \_\_init\_\_(self, val): \textbackslash{}n         self.val = val \textbackslash{}n         self.next =} \;$\cdot$\; \emph{from transformers import AutoModelForCausalLM \textbackslash{}n model = AutoModelForCausalLM.from\_pretrained(} \;$\cdot$\; \emph{async def fetch\_data(url): \textbackslash{}n     async with aiohttp.ClientSession() as} \;$\cdot$\; \emph{df = pd.DataFrame(\{'name': ['Alice', 'Bob'], 'age': [25, 30]\}) \textbackslash{}n df.groupby(} \;$\cdot$\; \emph{SELECT * FROM users WHERE} \;$\cdot$\; \emph{SELECT COUNT(*) FROM orders GROUP BY} \;$\cdot$\; \emph{CREATE TABLE employees ( \textbackslash{}n     id INTEGER PRIMARY KEY, \textbackslash{}n     name} \;$\cdot$\; \emph{const fetchData = async (url) =\textgreater{} \{ \textbackslash{}n     const response = await} \;$\cdot$\; \emph{document.addEventListener('DOMContentLoaded', () =\textgreater{} \{ \textbackslash{}n     const} \;$\cdot$\; \emph{const arr = [3, 1, 4, 1, 5]; \textbackslash{}n arr.sort((a, b) =\textgreater{}} \;$\cdot$\; \emph{fn main() \{ \textbackslash{}n     let mut v: Vec\textless{}i32\textgreater{} = Vec::new(); \textbackslash{}n     v.push(} \;$\cdot$\; \emph{\#include \textless{}iostream\textgreater{} \textbackslash{}n int main() \{ \textbackslash{}n     std::cout \textless{}\textless{}} \;$\cdot$\; \emph{\#!/bin/bash \textbackslash{}n for file in *.txt; do \textbackslash{}n     echo} \;$\cdot$\; \emph{The time complexity of binary search is} \;$\cdot$\; \emph{The time complexity of quicksort in the average case is} \;$\cdot$\; \emph{In Python, a decorator is defined using the} \;$\cdot$\; \emph{The difference between a stack and a queue is} \;$\cdot$\; \emph{A hash table resolves collisions using} \;$\cdot$\; \emph{The CAP theorem states that a distributed system cannot simultaneously guarantee} \;$\cdot$\; \emph{In object-oriented programming, polymorphism means} \;$\cdot$\; \emph{The SOLID principle 'S' stands for} \;$\cdot$\; \emph{A mutex differs from a semaphore in that} \;$\cdot$\; \emph{The purpose of a garbage collector is to} \;$\cdot$\; \emph{def divide(a, b): \textbackslash{}n     return a / b \textbackslash{}n \# potential issue:} \;$\cdot$\; \emph{def average(nums): \textbackslash{}n     return sum(nums) / len(nums) \textbackslash{}n \# potential issue:} \;$\cdot$\; \emph{int arr[10]; \textbackslash{}n for(int i=0; i\textless{}=10; i++) arr[i] =} \;$\cdot$\; 
\paragraph{History (common).}
\emph{The first president of the United States was} \;$\cdot$\; \emph{World War II ended in the year} \;$\cdot$\; \emph{The Berlin Wall fell in} \;$\cdot$\; \emph{The French Revolution began in} \;$\cdot$\; \emph{The first man to walk on the Moon was} \;$\cdot$\; \emph{The year Christopher Columbus reached the Americas was} \;$\cdot$\; \emph{The Declaration of Independence was signed in} \;$\cdot$\; \emph{The Renaissance began in} \;$\cdot$\; \emph{The inventor of the telephone was} \;$\cdot$\; \emph{The first emperor of Rome was} \;$\cdot$\; \emph{The year the Titanic sank was} \;$\cdot$\; \emph{The ancient civilization that built the pyramids of Giza was} \;$\cdot$\; \emph{The printing press was invented by} \;$\cdot$\; 
\paragraph{History (niche).}
\emph{The Treaty of Westphalia was signed in} \;$\cdot$\; \emph{The last emperor of the Byzantine Empire was} \;$\cdot$\; \emph{The Taiping Rebellion was led by} \;$\cdot$\; \emph{The Congress of Vienna took place in} \;$\cdot$\; \emph{The Battle of Thermopylae was fought between} \;$\cdot$\; \emph{The Meiji Restoration began in} \;$\cdot$\; \emph{The Edict of Nantes was issued by} \;$\cdot$\; \emph{The founder of the Mongol Empire was} \;$\cdot$\; \emph{The War of the Roses was fought between the houses of} \;$\cdot$\; \emph{The first Shogun of the Tokugawa shogunate was} \;$\cdot$\; \emph{The ancient city of Carthage was located in modern-day} \;$\cdot$\; \emph{The Rosetta Stone was discovered in} \;$\cdot$\; 
\paragraph{Math / Reasoning.}
\emph{The square root of 144 is} \;$\cdot$\; \emph{The value of pi to five decimal places is} \;$\cdot$\; \emph{What is 7 times 8? The answer is} \;$\cdot$\; \emph{What is 13 squared? The answer is} \;$\cdot$\; \emph{The factorial of 6 is} \;$\cdot$\; \emph{The integral of e\^{}x dx is} \;$\cdot$\; \emph{The derivative of sin(x) is} \;$\cdot$\; \emph{The derivative of ln(x) is} \;$\cdot$\; \emph{The determinant of a 2x2 matrix [[a,b],[c,d]] is} \;$\cdot$\; \emph{The eigenvalues of the identity matrix are} \;$\cdot$\; \emph{Euler's identity states that e\^{}(i*pi) + 1 =} \;$\cdot$\; \emph{The Pythagorean theorem states that} \;$\cdot$\; \emph{The fundamental theorem of calculus connects} \;$\cdot$\; \emph{A prime number is a number that} \;$\cdot$\; \emph{The Fibonacci sequence starts with} \;$\cdot$\; \emph{The number of legs on the animal that spins webs is} \;$\cdot$\; \emph{The language spoken in the country where the Eiffel Tower is located is} \;$\cdot$\; \emph{The currency used in the land of the rising sun is} \;$\cdot$\; \emph{The color you get when you mix red and blue is} \;$\cdot$\; \emph{The number of sides on a shape called a hexagon is} \;$\cdot$\; \emph{If a train travels at 60 mph for 2 hours, it covers} \;$\cdot$\; \emph{The planet known as the Red Planet is} \;$\cdot$\; \emph{The metal with the chemical symbol Fe is} \;$\cdot$\; \emph{The organ that pumps blood through the body is the} \;$\cdot$\; \emph{The gas that plants absorb from the atmosphere is} \;$\cdot$\; 
\paragraph{Culture / Language.}
\emph{The word 'Schadenfreude' means} \;$\cdot$\; \emph{The Japanese word 'tsunami' literally means} \;$\cdot$\; \emph{The Arabic word 'inshallah' translates to} \;$\cdot$\; \emph{The Hindi word 'namaste' means} \;$\cdot$\; \emph{The Latin phrase 'carpe diem' means} \;$\cdot$\; \emph{The French word for 'butterfly' is} \;$\cdot$\; \emph{The author of 'One Hundred Years of Solitude' is} \;$\cdot$\; \emph{The protagonist of 'Crime and Punishment' is} \;$\cdot$\; \emph{The author of 'The Tale of Genji' is} \;$\cdot$\; \emph{The author of 'Don Quixote' is} \;$\cdot$\; \emph{In Greek mythology, the god of the underworld is} \;$\cdot$\; \emph{In Norse mythology, the world tree is called} \;$\cdot$\; \emph{In Hindu mythology, the god of destruction is} \;$\cdot$\; \emph{The composer of 'The Four Seasons' is} \;$\cdot$\; \emph{The number of symphonies composed by Beethoven is} \;$\cdot$\; \emph{The painter of 'Starry Night' is} \;$\cdot$\; \emph{The sculptor of 'David' in Florence is} \;$\cdot$\; \emph{The Japanese dish made of vinegared rice and raw fish is called} \;$\cdot$\; \emph{The country where cricket originated is} \;$\cdot$\; \emph{The holy book of Islam is called} \;$\cdot$\; 
\paragraph{Chinese-specific.}
\begin{CJK}{UTF8}{gbsn}\emph{中国最长的河流是}\end{CJK} \;$\cdot$\; \begin{CJK}{UTF8}{gbsn}\emph{中国面积最大的省份是}\end{CJK} \;$\cdot$\; \begin{CJK}{UTF8}{gbsn}\emph{长城的东端起点是}\end{CJK} \;$\cdot$\; \begin{CJK}{UTF8}{gbsn}\emph{中国五岳中最高的山是}\end{CJK} \;$\cdot$\; \begin{CJK}{UTF8}{gbsn}\emph{秦始皇统一六国的年份是}\end{CJK} \;$\cdot$\; \begin{CJK}{UTF8}{gbsn}\emph{唐朝的都城是}\end{CJK} \;$\cdot$\; \begin{CJK}{UTF8}{gbsn}\emph{清朝的最后一位皇帝是}\end{CJK} \;$\cdot$\; \begin{CJK}{UTF8}{gbsn}\emph{四大发明包括造纸术、火药、印刷术和}\end{CJK} \;$\cdot$\; \begin{CJK}{UTF8}{gbsn}\emph{郑和下西洋开始于}\end{CJK} \;$\cdot$\; \begin{CJK}{UTF8}{gbsn}\emph{红楼梦的作者是}\end{CJK} \;$\cdot$\; \begin{CJK}{UTF8}{gbsn}\emph{三国演义中的三国是魏、蜀和}\end{CJK} \;$\cdot$\; \begin{CJK}{UTF8}{gbsn}\emph{水浒传中一共有多少位好汉}\end{CJK} \;$\cdot$\; \begin{CJK}{UTF8}{gbsn}\emph{李白最著名的诗之一是}\end{CJK} \;$\cdot$\; \begin{CJK}{UTF8}{gbsn}\emph{中国第一颗人造卫星的名字是}\end{CJK} \;$\cdot$\; \begin{CJK}{UTF8}{gbsn}\emph{青蒿素的发现者是}\end{CJK} \;$\cdot$\; \begin{CJK}{UTF8}{gbsn}\emph{中国传统节日中秋节吃的食物是}\end{CJK} \;$\cdot$\; \begin{CJK}{UTF8}{gbsn}\emph{中国象棋中将帅不能}\end{CJK} \;$\cdot$\; \begin{CJK}{UTF8}{gbsn}\emph{京剧中的四大行当是生旦净}\end{CJK} \;$\cdot$\; \begin{CJK}{UTF8}{gbsn}\emph{太极拳的基本理念来自}\end{CJK} \;$\cdot$\; \begin{CJK}{UTF8}{gbsn}\emph{台湾海峡连接的两个海域是}\end{CJK} \;$\cdot$\; 
\paragraph{Multilingual.}
\emph{La capitale de l'Allemagne est} \;$\cdot$\; \emph{Le plus grand fleuve de France est} \;$\cdot$\; \emph{L'auteur de 'Les Misérables' est} \;$\cdot$\; \emph{La tour Eiffel a été construite en} \;$\cdot$\; \emph{Die Hauptstadt von Österreich ist} \;$\cdot$\; \emph{Der höchste Berg Deutschlands ist} \;$\cdot$\; \emph{Die Formel für die kinetische Energie ist} \;$\cdot$\; \emph{Der Komponist der 'Mondscheinsonate' ist} \;$\cdot$\; \begin{CJK}{UTF8}{gbsn}\emph{日本で一番高い山は}\end{CJK} \;$\cdot$\; \begin{CJK}{UTF8}{gbsn}\emph{日本の首都は}\end{CJK} \;$\cdot$\; \begin{CJK}{UTF8}{gbsn}\emph{源氏物語の作者は}\end{CJK} \;$\cdot$\; \begin{CJK}{UTF8}{gbsn}\emph{日本で一番長い川は}\end{CJK} \;$\cdot$\;  \emph{La capital de Argentina es} \;$\cdot$\; \emph{El autor de 'Cien años de soledad' es} \;$\cdot$\; \begin{CJK}{UTF8}{mj}\emph{대한민국의 수도는}\end{CJK} \;$\cdot$\; \begin{CJK}{UTF8}{mj}\emph{한국에서 가장 높은 산은}\end{CJK} \;$\cdot$\; \emph{A capital do Brasil é} \;$\cdot$\; 
\paragraph{Temporal / Contemporary.}
\emph{The CEO of OpenAI is} \;$\cdot$\; \emph{The programming language Rust was created by} \;$\cdot$\; \emph{The transformer architecture was introduced in the paper titled} \;$\cdot$\; \emph{Bitcoin was created by} \;$\cdot$\; \emph{The company that created ChatGPT is} \;$\cdot$\; \emph{The founder of Tesla is} \;$\cdot$\; \emph{The population of Earth in 2023 was approximately} \;$\cdot$\; \emph{The tallest building in the world as of 2024 is} \;$\cdot$\; \emph{The latest major version of Python is} \;$\cdot$\; \emph{The host city of the 2024 Summer Olympics was} \;$\cdot$\; \emph{The current Secretary-General of the United Nations is} \;$\cdot$\; \emph{The most recent country to join the European Union is} \;$\cdot$\; \emph{The AI model GPT-4 was released by} \;$\cdot$\; \emph{The company that developed the BERT model is} \;$\cdot$\; \emph{The deep learning framework PyTorch was developed by} \;$\cdot$\; 
\paragraph{Long-tail / Obscure.}
\emph{The national animal of Scotland is} \;$\cdot$\; \emph{The only letter not appearing in any US state name is} \;$\cdot$\; \emph{The shortest war in history lasted approximately} \;$\cdot$\; \emph{The country with the most official languages is} \;$\cdot$\; \emph{The element with the highest melting point is} \;$\cdot$\; \emph{The first computer programmer is generally considered to be} \;$\cdot$\; \emph{The blood type known as the universal donor is} \;$\cdot$\; \emph{The language with the most native speakers in the world is} \;$\cdot$\; \emph{The phobia of long words is called} \;$\cdot$\; \emph{The only planet that rotates clockwise is} \;$\cdot$\; \emph{The inventor of the World Wide Web is} \;$\cdot$\; \emph{The hardest natural substance on Earth is} \;$\cdot$\; \emph{The tallest animal on Earth is} \;$\cdot$\; \emph{The only mammal capable of true flight is} \;$\cdot$\; \emph{The currency of Switzerland is} \;$\cdot$\; \emph{The SI unit of luminous intensity is} \;$\cdot$\; \emph{The Greek letter used to represent the golden ratio is} \;$\cdot$\; \emph{The number of bones in the adult human body is} \;$\cdot$\; \emph{The year the Internet was first made available to the public is} \;$\cdot$\; \emph{The astronomical unit is defined as the distance from} \;$\cdot$\; 
}